\documentclass[sigconf]{acmart}

\usepackage{booktabs} 
\usepackage{tabularray}
\usepackage{multirow}

\usepackage{tabularray}
\usepackage{graphicx}
\usepackage{subcaption}
\usepackage[ruled]{algorithm2e} 

\SetAlFnt{\small}
\SetAlCapFnt{\small}
\SetAlCapNameFnt{\small}
\SetAlCapHSkip{0pt}
\setcopyright{none}
\renewcommand\footnotetextcopyrightpermission[1]{}
\authorsaddresses{}
\usepackage[T1]{fontenc}
\DeclareRobustCommand{\augiefamily}{%
  \fontfamily{augie}\fontseries{m}\fontshape{n}\selectfont}
\DeclareTextFontCommand{\textaugie}{\augiefamily}

\copyrightyear{2024}
\acmYear{2024}
\setcopyright{acmlicensed}\acmConference[SIGGRAPH Conference Papers '24]{Special Interest Group on Computer Graphics and Interactive Techniques Conference Conference Papers '24}{July 27-August 1, 2024}{Denver, CO, USA}
\acmBooktitle{Special Interest Group on Computer Graphics and Interactive Techniques Conference Conference Papers '24 (SIGGRAPH Conference Papers '24), July 27-August 1, 2024, Denver, CO, USA}
\acmDOI{10.1145/3641519.3657504}
\acmISBN{979-8-4007-0525-0/24/07}
\acmJournal{TOG}

\newcommand{\tocite}[1]{\textcolor{red}{[TOCITE]}}

\begin{document}
\title{MaPa: Text-driven Photorealistic Material Painting for 3D Shapes}

\author{Shangzhan Zhang$^{1,2}$ \quad Sida Peng$^1$ \quad Tao Xu$^1$ \quad Yuanbo Yang$^1$ \quad Tianrun Chen$^1$ \\
Nan Xue$^2$ \quad Yujun Shen$^2$ \quad Hujun Bao$^{1}$ \quad Ruizhen Hu$^{3}$
\quad Xiaowei Zhou$^{1}$}
\affiliation{
\institution{$^1$Zhejiang University \quad $^2$Ant Group \quad $^3$Shenzhen University}
\country{}
}









\renewcommand\shortauthors{Zhang, S. et al}

%
%


%
%






\begin{teaserfigure}
\centering
\includegraphics[width=\textwidth]{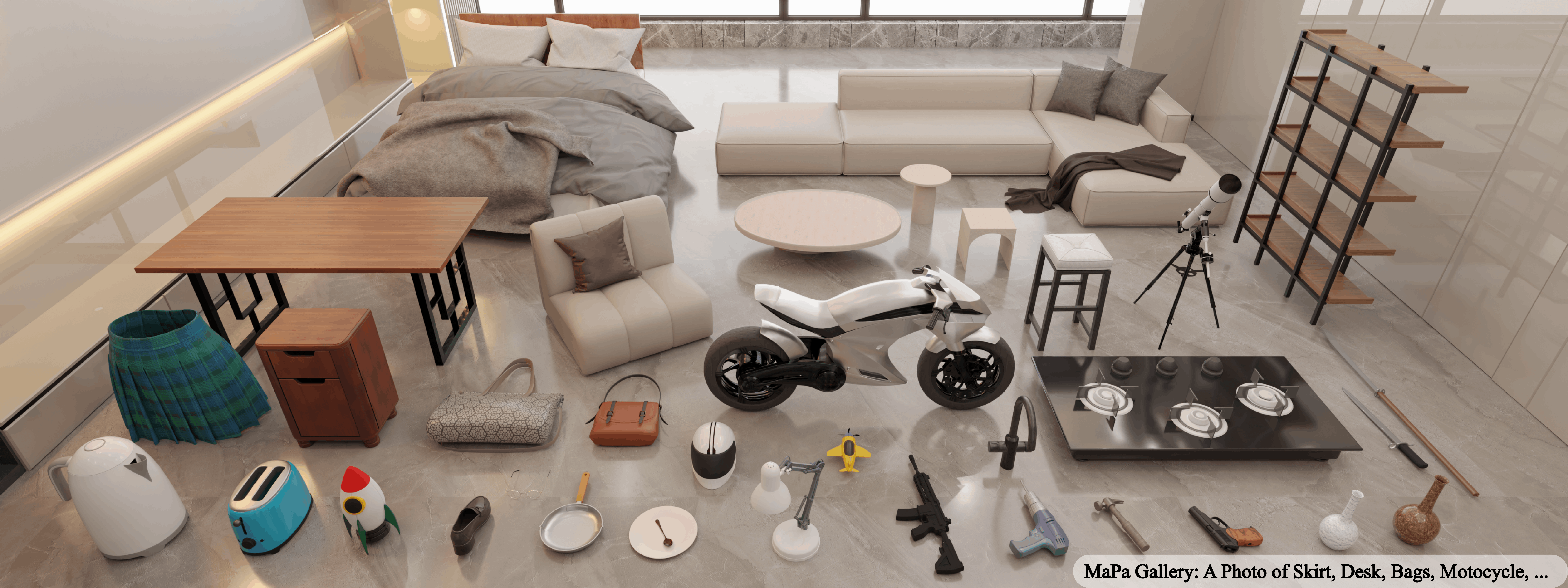}
\caption{
    Examples from \textbf{MaPa Gallery}, which facilitates photo-realistic 3D rendering by generating materials for daily objects.
    %
}
\vspace{3em}
\label{teaser} 
\end{teaserfigure}

\begin{abstract}

This paper aims to generate materials for 3D meshes from text descriptions. 
Unlike existing methods that synthesize texture maps, we propose to generate segment-wise procedural material graphs as the appearance representation, which supports high-quality rendering and provides substantial flexibility in editing. 
Instead of relying on extensive paired data, \textit{i.e.}, 3D meshes with material graphs and corresponding text descriptions, to train a material graph generative model, we propose to leverage the pre-trained 2D diffusion model as a bridge to connect the text and material graphs.
Specifically, our approach decomposes a shape into a set of segments and designs a segment-controlled diffusion model to synthesize 2D images that are aligned with mesh parts.
Based on generated images, we initialize parameters of material graphs and fine-tune them through the differentiable rendering module to produce materials in accordance with the textual description. 
Extensive experiments demonstrate the superior performance of our framework in photorealism, resolution, and editability over existing methods. Project page: \url{https://zhanghe3z.github.io/MaPa/}.

\end{abstract}

\maketitle

\section{Introduction}
3D content generation has attracted increasing attention due to its wide applications in video games, movies and VR/AR.
With the integration of diffusion models~\cite{rombach2022high, ho2020denoising, song2020denoising} into 3D generation~\cite{poole2022dreamfusion, liu2023zero1to3, cheng2023sdfusion}, this field has witnessed remarkable advancements and impressive results. 
Many works focus on generating 3D objects~\cite{poole2022dreamfusion, lin2023magic3d, liu2023zero1to3, long2023wonder3d} from text prompts or a single image.
In addition to 3D shape generation, generating appearance for existing meshes is also an important problem.
The conventional process of designing appearances for meshes involves extensive and laborious manual work, creating a pressing need within the community for a more efficient approach to producing mesh appearance.
Many 3D texture synthesis methods~\cite{Yeh2024TextureDreamerIT, richardson2023texture, NEURIPS2022_c7b925e6, cao2023texfusion, chen2023text2tex, metzer2022latent, yu2023texture} have been proposed to solve this problem, which can create diverse 3D textures for meshes according to the text input by users.
%

However, creating textures alone doesn't fully meet the needs of downstream applications, as we often render meshes under various lighting conditions. 
For example, in video games, choosing the right materials for rendering meshes in a scene is critical, as this impacts how the mesh interacts with the light around it, which in turn affects its appearance and realism. 
Therefore, designing an algorithm for painting materials on meshes is important.
Despite the remarkable success in texture generation, there is a lack of research focusing on generating high-quality materials for 3D objects.
Recently, Fantasia3D~\cite{Chen_2023_ICCV} tries to generate 3D objects with materials by distilling the appearance prior from 2D generative models. 
However, this approach faces substantial obstacles due to optimization instability and often fails to yield high-quality materials. 
Moreover, it models the object material as a per-point representation, which could be inconvenient for downstream user modifications.

In this paper, our goal is to generate photorealistic and high-resolution materials for meshes from textual descriptions, which can be conveniently edited by users.
We observe that the materials of newly manufactured objects in real life often exhibit consistency within specific areas due to the nature of the manufacturing processes.
Motivated by this, we hope that our generation process can mimic this characteristic, which has the advantage of leading to more organized results and allowing users to effortlessly swap the material of a specific area in modeling software, without the need to recreate the entire material map.
However, it is still an unsolved problem to separately generate materials for each part of the mesh and ensure the consistency of materials within the local region.

To this end, we propose a novel framework, named MaPa, to generate materials for a given mesh based on text prompts.
Our key idea is to introduce a segment-wise procedural material graphs representation for text-driven material painting and leverage 2D images as a bridge to connect the text and materials.
Procedural material graphs~\cite{li2023end, shi2020match, hu2022inverse}, a staple in the computer graphics industry, consist of a range of simple image processing operations with a set of parameters. They are renowned for their high-quality output and resolution independence, and provide substantial flexibility in editing.  
The segment-wise representation mimics the consistency of materials in the real world, making the results look more realistic and clean.
However, to train a generative model that directly creates material graphs for an input mesh, extensive data for pairs of texts and meshes with material graphs need to be collected.
 Instead, we leverage the pre-trained 2D diffusion model as an intermediate bridge to guide the optimization of procedural material graphs and produce diverse material for given meshes according to textual descriptions.


Specifically, our method contains two main components: segment-controlled image generation and material graph optimization.
Given a mesh, it is firstly segmented into several segments and projected onto a particular viewpoint to produce a 2D segmentation image.
Then, we design a segment-controlled diffusion model to generate an RGB image from the 2D segmentation.
In contrast to holistically synthesizing images~\cite{richardson2023texture}, conditioning the diffusion model on the projections of 3D segments can create 2D images that more accurately align with parts of the mesh, thereby enhancing the stability of the subsequent optimization process.
Subsequently, we produce the object material by estimating segment-wise material graphs from the generated image.
The material graphs are initialized by retrieving the most similar ones from our collected material graph library, and then their parameters are optimized to fit the image through a differentiable rendering module.
The generated materials can then be imported into existing commercial software for users to edit and generate various material variants conveniently.

We extensively validate the effectiveness of our method on different categories of objects. 
Our results outperform three strong baselines in the task of text-driven appearance modeling, in terms of FID and KID metrics, as well as in user study evaluation.  We also provide qualitative comparisons to the baselines, showcasing our superior photorealistic visual quality.
Users can easily use our method to generate high-quality materials for input meshes through text and edit them conveniently. Meanwhile, since we can generate arbitrary-resolution and tileable material maps, we can render fine material details at high resolution. 
\section{Related Works}
\paragraph{2D diffusion models.}
Recently, diffusion models~\cite{nichol2021glide, rombach2022high, ho2020denoising, song2020denoising} have been dominating the field of image generation and editing.
These works can generate images from text prompts, achieving impressive results.
Among them, Stable Diffusion~\cite{rombach2022high} is a large-scale diffusion model that is widely used.
ControlNet~\cite{zhang2023adding} introduces spatial conditioning controls (e.g., inpainting masks, Canny edges~\cite{canny1986computational}, depth maps) to pretrained text-to-image diffusion models. 
These signals provide more precise control over the diffusion process.
Another line of work focuses on controlling diffusion models with exemplar images.
Textual Inversion~\cite{gal2022textual} represents the exemplar images as learned tokens in textual space.
DreamBooth~\cite{ruiz2023dreambooth} finetunes the diffusion model and uses several tricks to prevent overfitting.
IP-Adapter~\cite{ye2023ip-adapter} trains a small adapter network to map the exemplar image to the latent space of the diffusion model.

\begin{figure*}[htbp]

\begin{center}
\includegraphics[width=\textwidth]{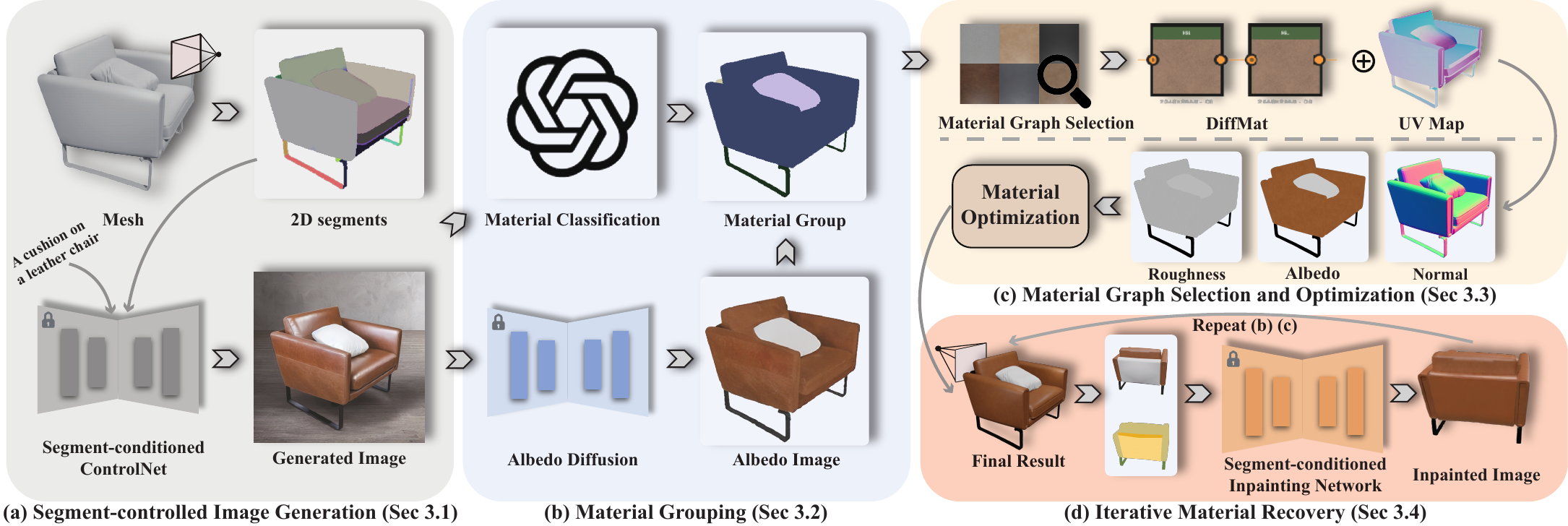}

\end{center}
\vspace{-1em}
\caption{
    \textbf{Illustration of our pipeline.} 
    Our pipeline primarily consists of four steps: a) \textbf{Segment-controlled image generation.} First, we decompose the input mesh into various segments, project these segments onto 2D images, and then generate the corresponding images using the segment-controlled ControlNet. b) \textbf{Material grouping.} We group segments that share the same material and have similar appearance into a material group. c) \textbf{Material graph selection and optimization.} For each material group, we select an appropriate material graph based on generated images and then optimize this material graph. d) \textbf{Iterative material recovery.} We render additional views of the input mesh with the optimized material graphs, inpaint the missing regions in these rendered images, and repeat steps b) and c) until all segments are assigned with material graphs.
}
\label{pipeline}
\vspace{-1em}

\end{figure*}


\paragraph{Appearance modeling.}
With the emergence of diffusion models, some works begin to use diffusion models to generate textures for 3D objects.
Latent-nerf~\cite{metzer2022latent} extends DreamFusion to optimize the texture map for a given 3D object.
However, Latent-nerf requires a long training time and produces low-quality texture maps.
TEXTure~\cite{richardson2023texture} and Text2tex~\cite{chen2023text2tex} utilize a pretrained depth-to-image diffusion model to iteratively paint a 3D model from different viewpoints.
TexFusion~\cite{cao2023texfusion} introduces an iterative aggregation scheme from different viewpoints during the denoising process.
These works greatly improve the quality of the generated texture map and speed up the  texture generation process.
However, these works can only generate texture maps, not realistic materials.
Material modeling~\cite{guarnera2016brdf} has long been a problem of interest in computer graphics and vision.
PhotoShape~\cite{park2018photoshape} uses shape collections, material collections, and image collections to learn material assignment to 3D shapes based on given images, where the shape and images need to be precisely aligned. 
TMT~\cite{hu2022photo} addresses the limitations of data constraints in PhotoShape by allowing different shape structures. However, its successful implementation necessitates a substantial quantity of 3D data with part segmentation~\cite{mo2019partnet} for training, which makes it can only work on a few categories of objects.
Moreover, both of those works treat material modeling as a classification problem and can only use materials from the pre-collected dataset, which often leads to incorrect material prediction under challenging lighting and dissimilar material assignment due to the restriction of dataset.
MATch~\cite{shi2020match} and its extension~\cite{li2023end} introduce a differentiable procedural material graphs. Procedural material graphs are popular in the graphics industry, which can generate resolution-invariant, tileable, and photorealistic materials.
PhotoScene~\cite{yeh2022photoscene} and PSDR-Room~\cite{yan2023psdr} produce the materials of indoor scenes from single images using differentiable procedural material graphs.
Different from these works, we focus on assigning realistic materials to a given object.
The closest work is Fantasia3D~\cite{Chen_2023_ICCV}, which follows DreamFusion and incorporates a spatially varying bidirectional reflectance distribution function (BRDF) to learn the surface materials for generated 3D objects.
However, the process of distilling 3D materials from a 2D pretrained diffusion model is not stable, and often fails to generate reasonable materials.
Moreover, the materials generated by Fantasia3D are often cartoonish and unrealistic.


\paragraph{Material and texture perception.}
Material perception and texture recognition are long-standing problems.
A considerable amount of earlier research on material recognition has concentrated on addressing the classification problem~\cite{liu2010exploring,bell2015material,hu2011toward,cimpoi2014deep}. 
Other works~\cite{sharma2023materialistic,upchurch2022dense,zheng2023materobot} attempt to solve material segmentation problem.
As for texture perception, a number of traditional approaches treat texture as a set of handcrafted features~\cite{caputo2005class, guo2010completed,lazebnik2005sparse}. 
Recently, deep learning methods~\cite{bruna2013invariant, chen2021deep, cimpoi2015deep, fujieda2017wavelet} have been introduced to address the problem of texture perception.
Our work benefits from these works, and we empirically find that GPT-4v~\cite{openai2023gpt} can be applied to material classification and texture description tasks in a zero-shot manner.

\section{Method}
Given a mesh and a textual prompt, our method aims to produce suitable materials to create photorealistic and relightable 3D shapes.
The overview of the proposed model is illustrated in Figure~\ref{pipeline}.
Section~\ref{sec:image_generation} first describes our proposed segment-controlled image generation.
The image-based material optimization is divided into three components: material grouping, material graph selection and optimization, and iterative material recovery.
Section~\ref{sec:Material_Grouping} provides details on using the generated image for material grouping.
Section~\ref{sec:material_recovery} introduces how to select initial material graphs from the material graph library and optimize the parameters of procedural node graphs according to the generated image.
Then, we design an iterative approach to generate materials for segments of the mesh that have not been assigned materials in Section~\ref{sec:iterative_material_recovery}.
Finally, we introduce a downstream editing module in Section~\ref{sec:downstream} to allow users to edit the generated materials conveniently.

\subsection{Segment-controlled image generation}
\label{sec:image_generation}
Given a mesh, we first oversegment it into a series of segments $\left\{\mathbf{s_i}\right\}_{i=1}^{N}$.
During the modeling process, designers often create individual parts, which are later assembled into a complete model. Hence, our method can separate these components by grouping the connected components. 
If the mesh is scanned or watertight, we can also use existing techniques~\cite{katz2003hierarchical} to decompose it into a series of segments.
Motivated by the progress of 2D generative models~\cite{zhang2023adding, rombach2022high, ho2020denoising}, we opt to project these segments to a viewpoint to generate 2D segmentation masks and perform subsequent conditional generation in 2D space.
For the good performance of conditional 2D generation, we carefully select a viewpoint to ensure a good initial result.
Specifically, a set of viewpoints are uniformly sampled with a 360-degree azimuth range, a constant 25-degree elevation angle, and a fixed radius of 2 units from the mesh.
We project the 3D segments onto these viewpoints and choose the one with the highest number of 2D segments as our starting viewpoint.
If multiple viewpoints have the same number of 2D segments, we empirically choose the viewpoint with the largest projected area of the mesh, as it is more likely to contain more details.
The 2D segments are sorted by area size and numbered from large to small to form the 2D segmentation mask.

From this viewpoint, the 2D segmentation mask is used to guide the generation of the corresponding RGB image, which then facilitates the process of material prediction that follows.
Note that previous works~\cite{richardson2023texture, chen2023text2tex, cao2023texfusion} usually use depth-to-image diffusion model to generate image conditioned on meshes, which may result in multiple color blocks within a segment and lead to unstable material optimization, thus we choose ControlNet with SAM mask~\cite{kirillov2023segment} as condition~\cite{gao2023editanything} for our segment-controlled image generation.
This process is defined as:
\begin{equation}
\label{eq:projection_generate}
\begin{aligned}
    \left\{\mathbf{\hat{s}_i}\right\}_{i=1}^{M} &= \mathcal{P}(\left\{\mathbf{s_i}\right\}_{i=1}^{N}), \\
    z_{\mathrm{t-1}} &= \boldsymbol{\epsilon}_\theta(z_{\mathrm{t}}, \boldsymbol{t}, \left\{\mathbf{\hat{s}_i}\right\}_{i=1}^{M}),
\end{aligned}
\end{equation}
where $\mathcal{P}$ is the projection function, and $\boldsymbol{\epsilon}_\theta$ is the pre-trained ControlNet. $z_{\mathrm{t}}$ is the noised latent at time step $t$, $\boldsymbol{t}$ is the time step, and $\left\{\mathbf{\hat{s}_i}\right\}_{i=1}^{M}$ is the 2D segments.  We finetune the SAM-conditioned ControlNet~\cite{gao2023editanything} on our own collected  dataset to obtain a segment-conditioned
ControlNet.
The implementation details of segment-conditioned
ControlNet and the comparison to depth-to-image diffusion model are presented in the supplementary material.

\subsection{Material grouping}
\label{sec:Material_Grouping}
To save the optimization time and obtain visually more coherent results, we first merge segments with similar appearance into groups and generate one material graph per group, instead of directly optimizing the material for each segment.
Specifically, we build a graph with each segment as a node, and connect two segments if they have the same material class and similar colors. Then, we extract all connected components from the graph as the grouping results.

In more details, for material classification, we use GPT-4v~\cite{openai2023gpt} due to its strong visual perception capabilities and convenient in-context learning abilities. We empirically find that GPT-4v works well with images created by a diffusion model. 
The details of the prompt are presented in the supplementary material.
For the color similarity, we train an albedo estimation network to remove effects caused by the shadows and strong lighting in the generated images.
We fine-tune Stable Diffusion conditioned on CLIP image embeddings on the ABO material dataset~\cite{collins2022abo}, with a RGB image as the input and the corresponding albedo image as the output.
The implementation details of albedo estimation network are presented in the supplementary material.
If the distance between the median color of two segments in the albedo image in the CIE color space~\cite{sharma2005ciede2000} is less than a threshold $\lambda$, these two segments are considered to be similar. We empirically set $\lambda$ to 2, which means the difference is perceptible through close observation~\cite{minaker2021optimizing}.

\subsection{Material graph selection and optimization}
\label{sec:material_recovery}
In this section, we describe how to recover the parameters of procedural node graphs from the generated images.
For each material group, we retrieve the most similar material graph from the material graph library. 
Then, we optimize the material graph to obtain the final material graph.

\paragraph{Material graph selection.}
We build a material graph library in advance, which contains a variety of 
material categories.
Thumbnails of material graphs are rendered using eight different environment maps to enhance the robustness of the retrieval process.
Following PSDR-Room~\cite{yan2023psdr}, we adopt CLIP for zero-shot retrieval of material graphs.
We crop the material group region from the generated image and compute its cosine similarity with same-class material graphs in our library through CLIP model.
The material graph with the highest similarity is selected. Similar to PhotoScene~\cite{yeh2022photoscene}, we apply homogeneous materials to material groups with an area smaller than 1000 pixels, as it is challenging to discern spatial variations in such small area.

\paragraph{Material graph optimization.}
To make the retrieved material graph close to the generated image, we optimize the parameters of the material graph using our differentiable rendering module.
The differentiable rendering module contains a differentiable renderer and DiffMat v2~\cite{li2023end}.
DiffMat v2~\cite{li2023end} is a framework that differentiably converts material graphs into texture-space maps, such as albedo map $\mathbf{A_{uv}}$, normal map $\mathbf{N_{uv}}$, roughness map $\mathbf{R_{uv}}$, etc.
We utilize UV maps generated by Blender to sample per-pixel material maps $\mathbf{A}$, $\mathbf{N'}$, $\mathbf{R}$. The normals $\mathbf{N'}$ should be rotated to align with the local shading frame to obtain the final normal map $\mathbf{N}$.
We use physically-based microfacet BRDF~\cite{karis2013real} as our material model. 
Our lighting model is 2D spatially-varying incoming lighting~\cite{li2020inverse}.
The incoming lighting is stored in a two-dimensional grid that varies spatially.
Using this lighting model significantly enhances the efficiency of our rendering process, which eliminates the need to trace additional rays.
InvRenderNet~\cite{li2020inverse} is a network that can predict the 2D spatially-varying lighting from a single image.
We use the prediction of InvRenderNet as our initial lighting $\mathbf{L_{init}}$.
During training, we optimize the lighting $\mathbf{L}$ by adding residual lighting $\mathbf{\widetilde{L}}$ to $\mathbf{L_{init}}$. 
The information of $\mathbf{\widetilde{L}}$ is stored in a 2D hash grid~\cite{mueller2022instant}.
For each pixel, we query the corresponding feature from the hash grid and use a shallow MLP to convert it into a residual lighting.
We set the initial value of $\mathbf{\widetilde{L}}$ to 0, and use the ReLU activation function to prevent the value of $\mathbf{L}$ from being negative.
The differentiable rendering module is described as:
\begin{equation}
\begin{aligned}
\label{eq:render}
\mathbf{A_{uv}}, \mathbf{N_{uv}}, \mathbf{R_{uv}} & =\operatorname{DiffMat}(\mathbf{G}), \\
\mathbf{A}, \mathbf{R} & = \operatorname{S}(\mathbf{A_{uv}},  \mathbf{R_{uv}}), \\
\mathbf{N} &= \operatorname{Rot}(\operatorname{S}(\mathbf{N_{uv}})), \\
\mathbf{I}_{rend} & = \mathcal{R}(\mathbf{A}, \mathbf{N}, \mathbf{R}, \mathbf{L}),
\end{aligned}
\end{equation}
where  $\mathbf{G}$ is the procedural material graph, $\operatorname{S}$ is the UV sampling function, $\operatorname{Rot}$ is the rotation function that rotates the normals $\mathbf{N'}$, and $\mathcal{R}$ is the differentiable renderer adopted from InvRenderNet~\cite{li2020inverse}. 
We can render a 2D image using the differentiable rendering module, and then optimize the parameters of the material graph by making the rendered image as similar as possible to the generated image.
However, images generated by the diffusion model typically have strong lighting and shadows.
To avoid overfitting, we introduce a regularization term that constrains the rendering result of the albedo map $\mathbf{A}$ of the material graph to be as similar as possible to the albedo image predicted by the network mentioned in Section~\ref{sec:Material_Grouping}.
Additionally, differentiable procedural material graphs highly constrain the optimization space of the material model, which can also alleviate the instability during the training process due to the presence of strong lighting and shadows.
The loss functions and more technical details are presented in the supplementary material.

\subsection{Iterative material recovery}
\label{sec:iterative_material_recovery}
The aforementioned process can only generate materials for one viewpoint of the mesh. 
For areas without material, we adopted an iterative approach to inpaint them.
Among all the viewpoints generated in Section~\ref{sec:image_generation}, we select the viewpoint adjacent to the initial one for the next iteration. 
During this iteration, it is necessary to determine which segments require material assignment. 
A segment is considered in need of material if it has not yet been assigned one, or if its projected area in the previous viewpoint is less than a quarter of its projected area in the current viewpoint. 

To ensure the consistency of the inpainting results, we adopt a common design~\cite{zeng2023paint3d, dihlmann2024signerf} in the community.
We render the mesh with partially assigned materials both from the previous iteration viewpoints and the current iteration viewpoint using Blender Cycles Renderer. 
Then, the previous iteration viewpoint images and the current iteration viewpoint image are concatenated as the input of the inpainting network.
Only the segments that need material assignment are inpainted, and the inpainting mask of other regions is set to 0.
The SAM-conditioned inpainting network~\cite{gao2023editanything} is adopted as our inpainting network.
We employ the process described in Section~\ref{sec:material_recovery} again to generate materials for these segments. 
If segments requiring material assignment and those already assigned material are grouped into the same material group, we directly copy the existing material graph to the segments requiring material.
Above steps are repeated until all regions are assigned materials.
\begin{figure}
\begin{center}
\newcommand{\mywidth}{\linewidth}
\includegraphics[width=\mywidth]{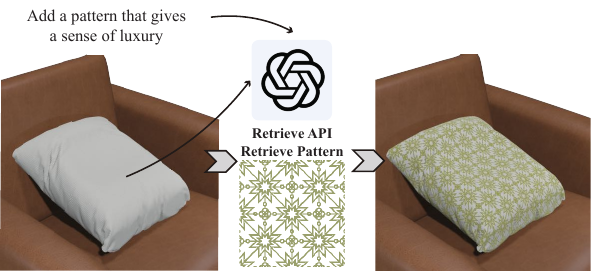}
\vspace{-2em}
\end{center}

\captionsetup{font={normalsize}} 
\caption{\textbf{Downstream editing.}
We perform material editing on generated material. The user can edit the material using textual prompts through the GPT-4 and a set of predifined APIs.
}
\label{fig:editing_method}
\end{figure}

\subsection{Downstream editing}
\label{sec:downstream}
Different from previous work~\cite{Chen_2023_ICCV, richardson2023texture, chen2023text2tex, cao2023texfusion}, our method can generate editable material graphs, which is paramount for users. 
Users can edit the material graph directly in Substance Designer~\cite{adobeDesignSoftware}.
For example, they can change the noise that affects the base color to change the texture, or add some nodes that affect the roughness to add details such as scratches and other surface imperfections.

However, it is difficult for most users to operate the underlying noise nodes which are complex and intertwined.
Inspired by the work of VISPROG~\cite{gupta2023visual}, we provide users with some high-level operations, allowing them to easily modify the material through textual instructions.
For example, when users input the requirement ``add a pattern that gives a sense of luxury'', our method will find a geometric seamless pattern mask that matches the description and generate the corresponding python command to add such pattern so that the user can run this command locally to obtain the desired material modification. 
One example result of adding texture is shown in Figure~\ref{fig:editing_method}. 
More details about this function can be found in the supplementary material.

\section{Experiments}

\subsection{Implementation details}
Our material graph optimization in three stages uses Adam~\cite{kingma2014adam} as the optimizer. We perform a total of 300 iterations for optimization. 
In general, the entire process of our method can be completed in less than 10 minutes on an A6000, which is faster than Text2tex~\cite{chen2023text2tex} (15 minutes) and Fantasia3D~\cite{Chen_2023_ICCV} (25 minutes).
We collect 8 material categories, including wood, metal, plastic, leather, fabric, stone, ceramic, and rubber. The number of each category is given in the supplementary material.
Since the eight materials mentioned above already cover the vast majority of man-made objects, we did not collect additional types of material graphs.

\subsection{Comparison to baselines}
\paragraph{Baselines.}
We compare our method with three strong baselines on text-driven appearance modeling.
TEXTure~\cite{richardson2023texture} introduces an iterative texture painting scheme to generate texture maps for 3D objects.
Text2tex~\cite{chen2023text2tex} uses a similar scheme to TEXTure. Additionally, it creates an automated strategy for selecting viewpoints during the iterative process.
Fantasia3D~\cite{Chen_2023_ICCV} separates the modeling of geometry and appearance, adopting an MLP to model the surface material properties of a 3D object.
For the following experiments, we employ an environmental map to generate outcomes for our method and Fantasia3D. The results for TEXTure and Text2tex are derived from their diffuse color pass in Blender.

\begin{table}\centering
    \caption{\textbf{Comparison results.} 
    We compare our method with three strong baselines. Our approach yields superior quantitative results and attains the highest ratings in user studies.
    }
    \vspace{-0.5em}
    \resizebox{\linewidth}{!}{
    \begin{tabular}{c c c c c c c}
    
    \toprule
    Dataset & Methods                    & FID$\downarrow$    & KID$\downarrow$  & Overall quality$\uparrow$ & Fidelity$\uparrow$ \\
    \midrule
    \multirow{4}{*}{\textbf{Chair}} & TEXTure     & 94.6 & 0.044 & 2.43 &  2.41  \\
    & Text2tex   & 102.1 & 0.048 & 2.35 &  2.51 \\
    & Fantasia3D   & 113.7  & 0.055 & 1.98 &   2.08  \\
    & \textbf{Ours}         & \textbf{88.3}    & \textbf{0.037}  & \textbf{4.33}    & \textbf{3.35}    \\

    \midrule
    \multirow{4}{*}{\textbf{ABO}} & TEXTure     & 118.0 & 0.027 & 2.65 &  2.41  \\
    &Text2tex   & 109.8 & 0.020 & 3.10 &  2.94 \\
    &Fantasia3D   & 138.1  & 0.034 & 1.51 &   1.80  \\
    &\textbf{Ours}         & \textbf{87.3}    & \textbf{0.014}  & \textbf{4.10}    & \textbf{3.22}  \\
    \bottomrule 
    \label{tab:quantitative}
    \end{tabular}}
    \label{usr}
    \vspace{-2em}
    
\end{table}

\paragraph{Quantitative comparison.}
Our goal is to obtain high rendering quality that is as close as possible to the real image, so we measure the distribution distance between the rendered image and the real image using FID~\cite{heusel2017gans} and KID~\cite{binkowski2018demystifying}.
We select 300 chairs from TMT~\cite{hu2022photo} and 30 random objects from ABO dataset~\cite{collins2022abo} as test data. 10 views are uniformly sampled for each chair with a 360-degree azimuth range, a constant 45-degree elevation angle, and a fixed radius of 1.5 units. 
The prompt in the experiment is extracted from the rendering of original objects using BLIP~\cite{li2022blip}. 
Each baseline uses the same prompts.
The segmentation of the each input shape is obtained by finding connected components.

Moreover, we also conducted a user study to assess the rendering results. 
Each respondent was asked to evaluate the results based on two aspects: overall quality and fidelity to the text prompt, using a scale of 1 to 5. 
Overall quality refers to the visual quality of the rendering results, while fidelity to the text prompt indicates how closely these results align with the given text prompt.
There were 40 users participated in the study, including 16 professional artists and 24 members of the general public. We gathered 60 responses from each participant, and the final score represents the average of all these responses.
As shown in Table~\ref{tab:quantitative}, our method achieves the best quantitative results and the highest user study scores.

\begin{figure}
\setlength\tabcolsep{0.2em}
\newcommand{\mywidth}{0.19 \textwidth}
\begin{center}
    \includegraphics[width=\linewidth]{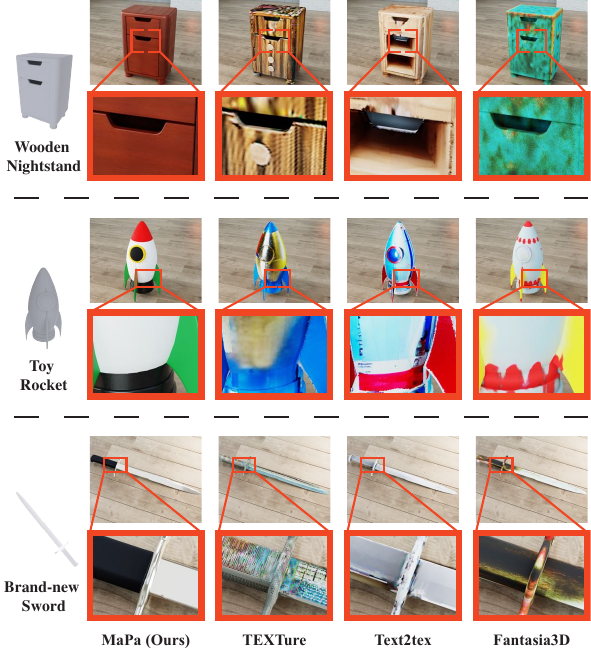}
\end{center}
\vspace{-1em}
\captionsetup{font={normalsize}} 
\caption{\textbf{Qualitative comparisons.} 
The results generated by our method and all the baselines are rendered in the same CG environment for comparison. The prompts for the three objects are: "a photo of a wooden bedside table," "a photo of a toy rocket," and "a photo of a brand-new sword."
}
\label{fig:qualitative}
\label{fig:exp}
\end{figure}

\paragraph{Qualitative comparison.}
We collect several high-quality 3D models from Sketchfab~\cite{sketchfabSketchfabBest} and use these data for qualitative experiments. 
As shown in Figure~\ref{fig:qualitative}, our method can achieve more realistic rendering results than the baselines.
TEXTure and Text2tex are prone to produce inconsistent textures, which makes their results look dirty.      
Fantasia3D's results frequently exhibit oversaturation or fail directly.

\subsection{More qualitative results}

\begin{figure}
\begin{center}
\includegraphics[width=\linewidth]{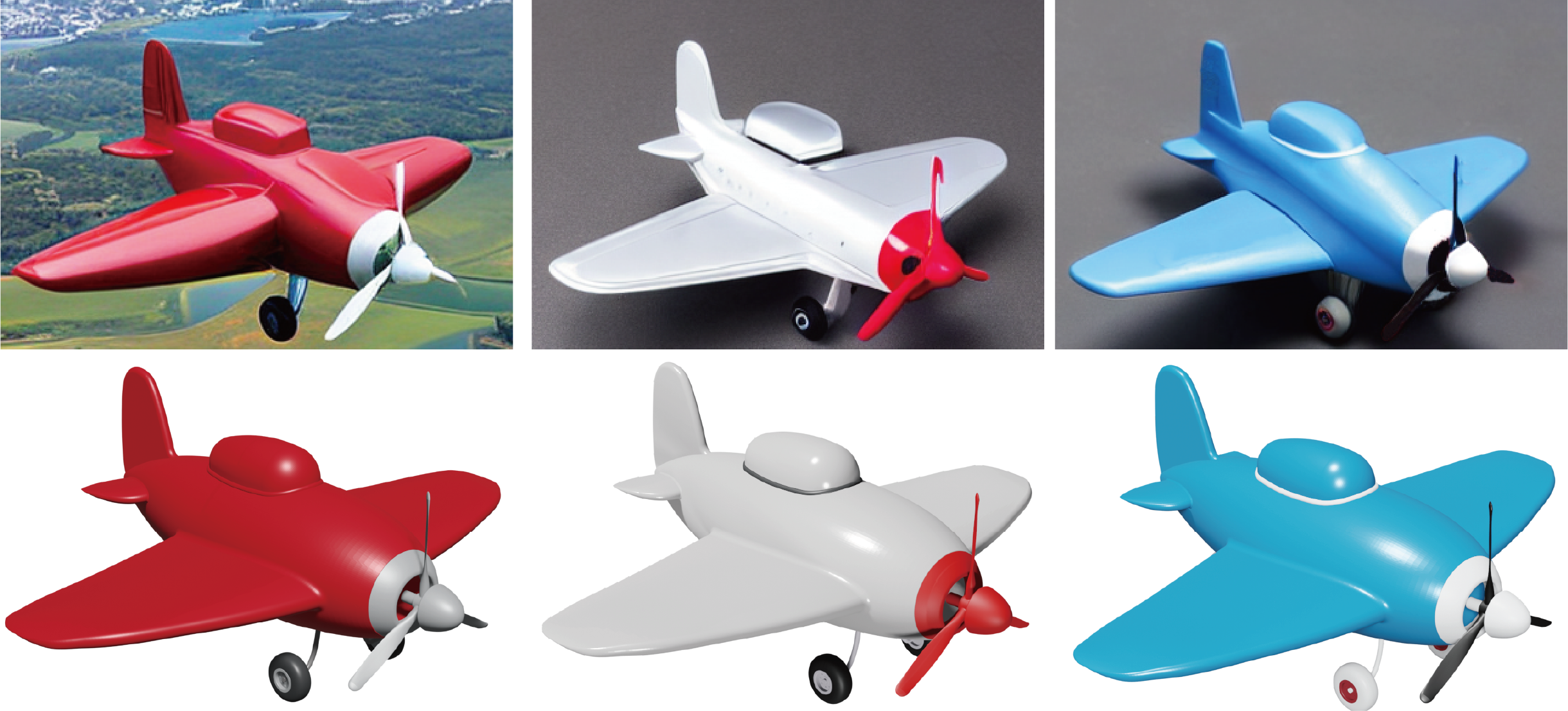}
\vspace{-2em}
\end{center}

\captionsetup{font={normalsize}} 
\caption{\textbf{Diversity of our generated material.}
We show the diversity of results synthesized by our framework with the same prompt: ``A photo of a toy airplane''.
Images in the first row are generated by diffusion models, and models in the second row are our painted meshes.
}
\label{fig:diversity}
\end{figure}

\paragraph{Diversity of results.}
2D diffusion model can generate different images with different seeds, which naturally brings diversity to our framework when optimizing material according to the generated images.
We show an example shape with diverse results in Figure~\ref{fig:diversity}. 
Note how the painted materials change when different images of the same object are generated.

\begin{figure}
\begin{center}
\includegraphics[width=\linewidth]{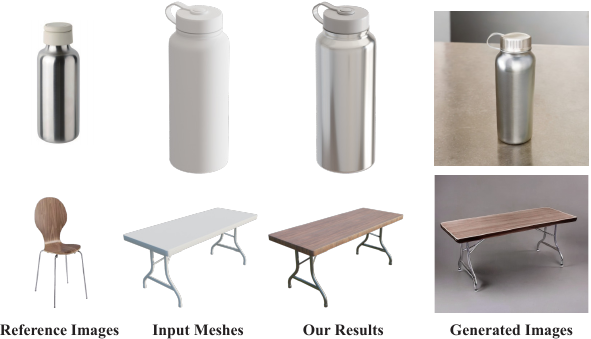}
\vspace{-2em}
\end{center}

\captionsetup{font={normalsize}} 
\caption{\textbf{Appearance transfer.}
Our method can also take image prompt as input, transferring the appearance of reference images to input objects.
}
\label{fig:reference}
\end{figure}

\paragraph{Appearance transfer from image prompt.}
We can use the image encoder introduced in IP-Adapter~\cite{ye2023ip-adapter} to be compatible with image prompt. 
As shown in Figure~\ref{fig:reference}, our method can transfer the appearance from the reference image to the input objects by first generating the corresponding image of the given object with similar appearance and then optimizing the materials using our framework.

\begin{figure*}
\vspace{-1em}
\begin{center}
\includegraphics[width=.79\linewidth]{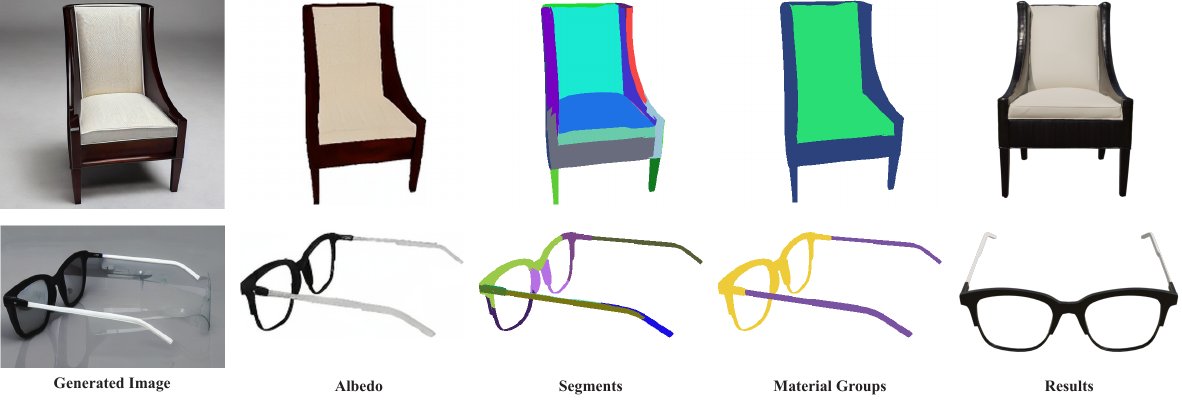}
\vspace{-1em}
\end{center}

\captionsetup{font={normalsize}} 
\caption{\textbf{Results on watertight meshes.}
We decompose the mesh into several segments using the graph cut algorithm and generate materials for each segment. The text prompts for those two examples are ``A photo of a modern chair with brown legs'' and "A black and white eyeglass", respectively.}
\label{fig:wtm}
\end{figure*}

\paragraph{Results on watertight meshes.}
Although we mainly focus on meshes that can be segmented by grouping connected components, our method can also handle watertight meshes. In Figure~\ref{fig:wtm}, we also show the results of painted watertight meshes. We use the graph cut algorithm~\cite{katz2003hierarchical} to segment the watertight mesh, and then we generate materials for each segment.

\begin{figure}
\begin{center}
\newcommand{\mywidth}{\linewidth}
\includegraphics[width=\mywidth]{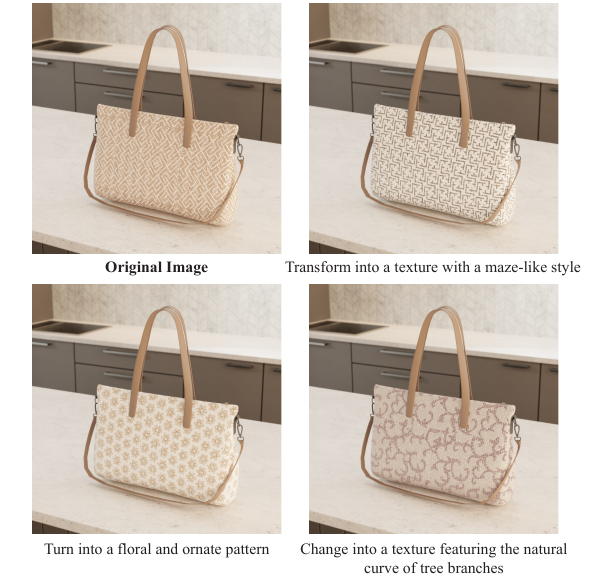}
\vspace{-1em}
\end{center}

\captionsetup{font={normalsize}} 
\caption{\textbf{Results of downstream editing.}
We perform text-driven editing on generated material through the GPT-4 and a set of predifined APIs. 
}
\label{fig:editing_bag}
\end{figure}

\paragraph{Downstream editing results.}
Our method can edit materials based on user-provided prompts, and we show several editing results on a bag in Figure~\ref{fig:editing_bag}. We can see that we are able to add different patterns to the bag according to the prompts.

\subsection{Ablation studies}

\begin{figure*}
\begin{center}
\vspace{-1em}
\includegraphics[width=.81\linewidth]
{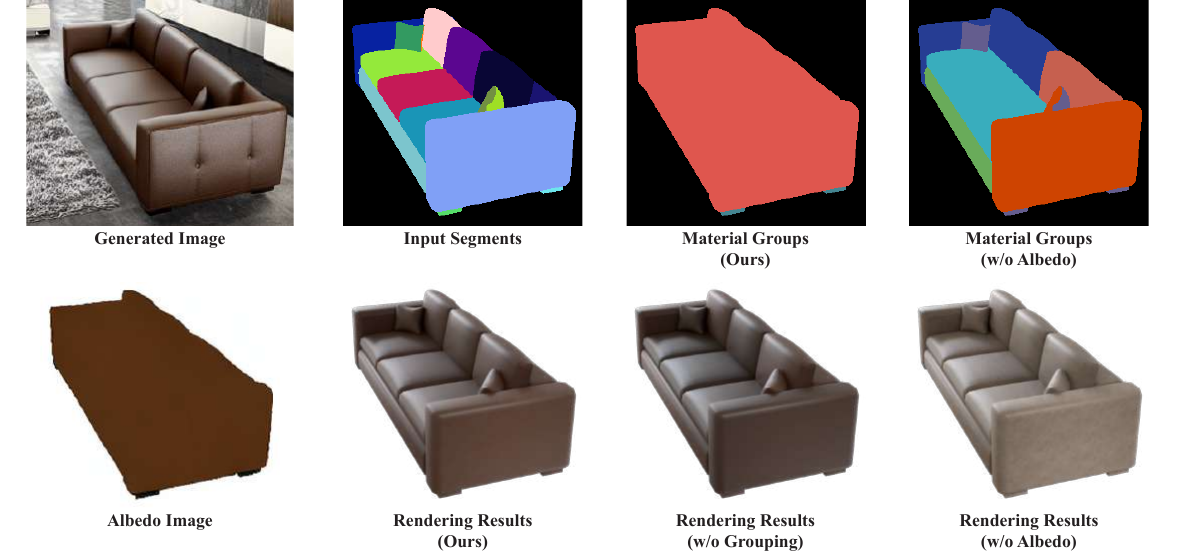}
\vspace{-1em}
\end{center}

\caption{\textbf{Ablation study of ``w/o Albedo'' and ``w/o Grouping''.}
Example with text prompt ``A dark brown leather sofa''.
``w/o Albedo'' exhibits color disparity between the rendering result and generated image, and fails to merge segments with similar color.
``w/o Grouping'' has similar rendering results to our method.
}
\label{fig:ablation_albedo}
\end{figure*}

\paragraph{The significance of albedo estimation network.}
To justify the usage of our albedo estimation network, we compare with the setting ``w/o Albedo'', where we remove this module from our method and use the generated image directly.
As illustrated in Figure~\ref{fig:ablation_albedo}, the baseline ``w/o Albedo'' fails to merge the segments with the similar color, hindered by the impacts of shadow and lighting.
Additionally, a noticeable color disparity exists between the rendering result of ``w/o Albedo'' and the generated image.

\paragraph{Robustness analysis of grouping results.}
Our method can also work without material grouping. We show the results of our method without material grouping in Figure~\ref{fig:ablation_albedo}, referred as ``w/o Grouping''. The results of ``w/o Grouping'' are not significantly different from our results. 
However, the optimization time will be much longer without grouping. 
We find our method takes 7 minutes on average to finish the all optimization process per shape. If we remove the material grouping, the optimization time on average
will increase to 33 minutes.

\paragraph{The significance of material graph optimization.}
We also show some representative visual comparisons with the setting ``w/o Opt'', where the material graph is selected without further optimization in Figure~\ref{fig:ablation_wood}.
We can clearly see that the optimization highly increases the appearance similarity to the generated images and is not limited by the small material graph data set.
Generated images (a), (b), and (c) are generated under the same pose. The retrieval results of the wood material graph in the three images are the same shown as ``w/o Opt''. This proves that even if our material graph data set is small, our method can still produce diverse texture results.
We also conduct an ablation study on effectiveness of residual lighting, refereed as ``w/o Res''.
Compared to our result (a), ``w/o Res'' produces textures that are inconsistent with the generated images. This is because the initial 2D spatially-varying lighting is not very accurate, which is caused by the domain gap between the training data of InvRenderNet and the images generated by the diffusion model. 



\section{Conclusions}
\begin{figure}
\begin{center}
\includegraphics[width=.88\linewidth]{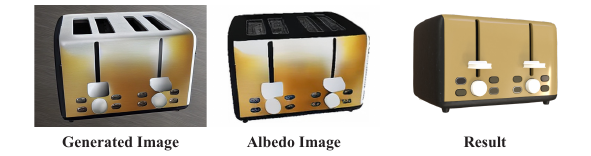}
\vspace{-1em}
\end{center}

\captionsetup{font={normalsize}} 
\caption{\textbf{Failure case.}
Because of the unusual light effects generated by diffusion (silver metallic with yellow light), the albedo estimation network fails to accurately estimate the albedo, leading to dissimilar material prediction.
}
\vspace{-1em}
\label{fig:failure}
\end{figure}

In this paper, we propose a novel method for generating high-quality material map for input 3D meshes. We introduce segment-wise procedural material graphs representation for this challenging task and adopt a pre-trained 2D diffusion model as an intermediary to guide the prediction of this representation. Our results yield photorealistic renderings that are also conveniently flexible for editing. It is  worth exploring ways to train an amortized inference network that can directly predict the parameters of the material graph from the input image to highly improve the efficiency. Our limitations are as follows:
\paragraph{Domain gap of generated images.
}
Our method sometimes fails to generate rendering results that match the generated image, as shown in Figure~\ref{fig:failure}, mainly due to the different domain of the diffusion model and our albedo estimation network. This is because the diffusion model is trained on natural images, while our albedo estimation network is trained on synthetic images. We suggest that fine-tuning the diffusion model on synthetic images could mitigate this issue. 
Alternatively, employing a more advanced albedo estimation network could potentially serve as a replacement for our current albedo estimation network. 
Moreover, we currently use GPT-4v for material classification as we found that its performance is better than other prior works, and it would be easy to be replaced if more powerful methods for material classification come out.

\begin{figure*}
\begin{center}
\vspace{-1em}
\includegraphics[width=.79\linewidth]{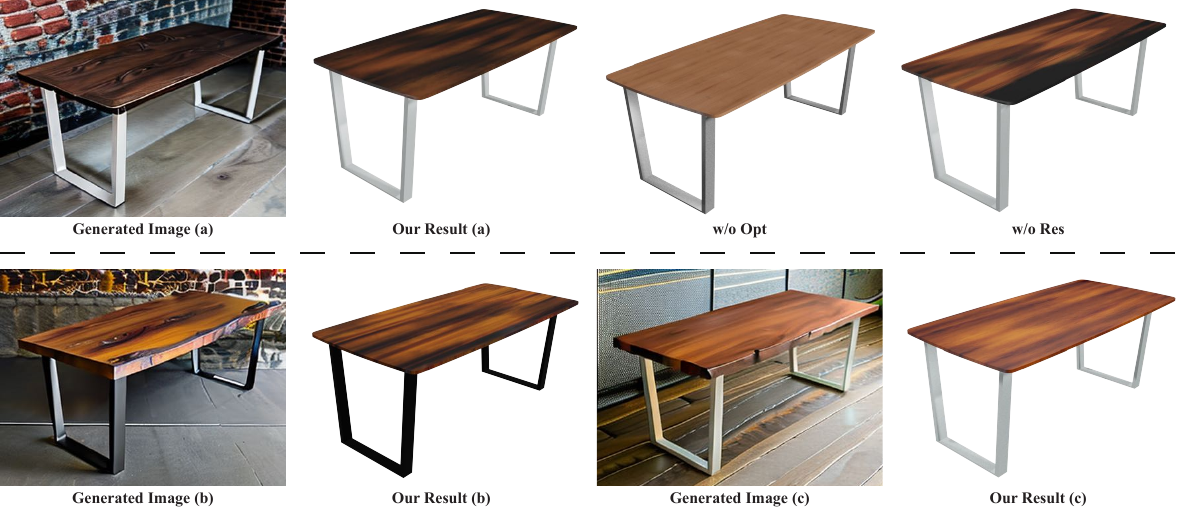}
\vspace{-1em}
\end{center}

\captionsetup{font={normalsize}} 
\caption{\textbf{Ablation study of our material graph optimization.}
We show the ablation study of our material graph optimization and the effectiveness of residual light.
The baseline ``w/o Opt'' and ``w/o Res'' are optimized to minimize the loss of the generated image (a) and the rendered image.
The generated image (b) and (c) are used to prove that our method can yield diverse results even with a small material graph dataset.
}
\label{fig:ablation_wood}
\end{figure*}

\paragraph{Complex objects with unobvious segments.}
For complex objects with unobvious segments, we tend to over-segment the shape and then group the segments with similar materials according to the generated image. Final results will depend on over-segmentations and the segment-conditioned ControlNet may fail to produce reasonable guidance for grouping due to the lack of training data of such complex objects.
\paragraph{Expressiveness of the material graph.}
The material can be optimized to produce spatially varying appearance only if the material graph contains some corresponding nodes for optimization, as the wood material shown in Figure~\ref{fig:ablation_wood}. Currently, most of the other materials we used lack such nodes, and the complex look in Figure~\ref{fig:editing_bag} indeed comes from the downstream editing by retrieving predefined patterns. We echo that the method could handle more complex targets if we have a larger collection of graphs with more expressiveness.

\begin{acks}
    We would like to thank Xiangyu Su for his valuable contributions to our discussions. This work was partially supported by NSFC (No. 62172364 and No. 62322207), Ant Group and Information Technology Center and State Key Lab of CAD\&CG, Zhejiang University.
\end{acks}
\bibliographystyle{ACM-Reference-Format}
\bibliography{ref}

\clearpage

\end{document}